# COMBINED INDEPENDENT COMPONENT ANALYSIS AND CANONICAL POLYADIC DECOMPOSITION VIA JOINT DIAGONALIZATION

*Xiao-Feng Gong, Cheng-Yuan Wang, Ya-Na Hao, and Qiu-Hua Lin*

Faculty of Electronic Information and Electrical Engineering
Dalian University of Technology, Dalian 116023, China
E-mail: xfgong@dlut.edu.cn

**ABSTRACT**

Recently, there has been a trend to combine independent component analysis and canonical polyadic decomposition (ICA-CPD) for an enhanced robustness for the computation of CPD, and ICA-CPD could be further converted into CPD of a 5th-order partially symmetric tensor, by calculating the eigenmatrices of the 4th-order cumulant slices of a trilinear mixture. In this study, we propose a new 5th-order CPD algorithm constrained with partial symmetry based on joint diagonalization. As the main steps involved in the proposed algorithm undergo no updating iterations for the loading matrices, it is much faster than the existing algorithm based on alternating least squares and enhanced line search, with competent performances. Simulation results are provided to demonstrate the performance of the proposed algorithm.

## 1. INTRODUCTION

The use of tensor tools for the analysis of multidimensional signals has attracted wide interests in the past decades. For example, tensors were used to formulize the multilinearities of higher-order statistics or nonstationary (or colored) 2nd order statistics in independent component analysis (ICA) [1-2]. They have also been used in practical systems where data acquisition is in nature multilinear and tensor model could be directly set up in the deterministic data domain [3-5]. The main merit of using tensors is the essential uniqueness of several tensor decomposition tools [6, 7]. In particular, the canonical polyadic decomposition (CPD, also known as canonical / parallel factor analysis: CPA) is among the most important tensorial tools.

Although CPD is essentially unique in theory, its actual computation is not always guaranteed to generate globally optimal results [8], and one way to address this problem is to incorporate extra priors into decomposition procedure [5, 9-11]. In particular, noting that it is often reasonable to assume statistical independence at one mode of the tensorial datasets, the idea of combining ICA and CPD emerged in biomedical applications [5], and was further developed in [10, 11]. More exactly, the methods in [5, 11] perform ICA firstly on the matricized tensor to extract independent components, and impose CPD structure afterwards via rank-1 approximation. As a contrary, the work in [10] proposes to incorporate CPD structure during the ICA computation, by converting the ICA-CPD problem into the CPD of a 5th-order partially symmetric tensor by calculating 4th-order cumulants of the 3-way datasets. In addition, a 5th-order partially symmetric CPD method is proposed based on alternating least squares (ALS) with enhanced line search (ELS) as the accelerator.

In this paper, we propose a new algorithm of 5th-order CPD with partial symmetry for ICA-CPD based on joint diagonalization (JD). More exactly, we matricize the target tensor and factorize it into the product of 2 matrices. Then we use rank-1 structure detector given in [12] on both these 2 matrices to link CPD constrained with partial symmetry to real-valued JD. Lastly, rank-1 approximation is used to obtain estimates of the loading matrices. We note that the proposed algorithm undergoes no updating iterations for the loading matrices and thus is expected to be computationally more efficient than the algorithm based on ALS and ELS [10].

In the rest of the paper, problem formulation is given in Section 2, and Section 3 presents the proposed algorithm. Simulation results are shown in Section 4, and Section 5 concludes this paper.

## 2. PROBLEM FORMULATION

We assume $R$ mutually independent non-Gaussian sources $s_1,\cdots,s_R \in \mathbb{C}^K$ are mixed with two sets of parallel loading factors $a_1,\cdots,a_R \in \mathbb{C}^I$, and $b_1,\cdots,b_R \in \mathbb{C}^J$:

$$\mathcal{X} = \text{Tri}(A, B, S) \triangleq \sum_{r=1}^{R} a_r \circ b_r \circ s_r \quad (1)$$

where "$\circ$" denotes tensor outer product, $\mathcal{X} \in \mathbb{C}^{I \times J \times K}$, $A \triangleq [a_1,\cdots,a_R]$, $B \triangleq [b_1,\cdots,b_R]$, and $S \triangleq [s_1,\cdots,s_R]$. We note

This work is supported by Doctoral Fund of Ministry of Education of China (No.20110041120019), and National Natural Science Foundation of China (Nos. 61072098, 61105008, 61331019, 61379012)

that (1) is actually a third-order CPD model with statistical independence constraint at the last mode.

We matricize $\mathcal{X}$ into a matrix $X \in \mathbb{C}^{IJ \times K}$ such that $X((i-1)J+j,k) = \mathcal{X}(i,j,k)$, and $X$ could be written as:

$$X = (A \odot B)S^T \quad (2)$$

where "$\odot$" is Khatri-Rao (column-wise Kronecker) product. In addition, by denoting $M = (A \odot B)$, we note that (2) actually infers a linear instantaneous mixing procedure of statistically independent sources, with the constraint that the mixing matrix is of Khatri-Rao structure.

To solve the above problem, [10] proposed to convert it into the CPD of a 5th-order tensor with partial symmetry, using 4th-order statistics. We summarize the main steps as:

- Calculate the sampled 4th-order cumulant matrix[1]:

$$C = \text{cum}(X, X^*, X^*, X) \in \mathbb{C}^{I^2J^2 \times I^2J^2} \quad (3)$$

- Perform eigenvalue decomposition (EVD) on $C$: $C = \sum_{r=1}^{I^2J^2} \lambda_r e_r e_r^H$, where $e_r$ is the eigenvector associated with the $r$th largest eigenvalue $\lambda_r$, and then calculate $R$ dominant eigenmatrices as follows:

$$E_r(i,j) = \sqrt{\lambda_r} e_r((i-1)IJ+j), \quad r = 1, 2, ..., R \quad (4)$$

where $i, j = 1, 2, ..., IJ$. According to [10], eigenmatrices $\{E_r, r=1,...,R\}$ are jointly diagonalizable if $S$ contain mutually independent columns, that is:

$$E_r = MD_rM^H \quad (5)$$

where $D_r = \text{diag}(d_{1,r}, d_{1,r}, ..., d_{R,r})$ is a diagonal matrix with $d_{l,r}$ being its $l$th diagonal entry, $l = 1,...,R$.

- Constructing a 5th-order tensor $\mathcal{T} \in \mathbb{C}^{I \times J \times I \times J \times R}$ as:

$$\mathcal{T}(i_1, j_1, i_2, j_2, r) = E_r((i_1-1)J+j_1, (i_2-1)J+j_2) \quad (6)$$

and denoting $d_r \triangleq [d_{1,r}, ..., d_{R,r}]^T$, we have:

$$\mathcal{T} = \sum_{r=1}^{R} a_r \circ b_r \circ a_r^* \circ b_r^* \circ d_r \quad (7)$$

Noting further that $C((i-1)IJ+j, r) = C^*((j-1)IJ+i, r)$, which yields $e_r((i-1)IJ+j) = e_r^*((j-1)IJ+i)$, we come to the conclusion that $E_r$ is Hermitian and thus $d_r$ is real-valued, $r = 1, 2,..., R$. As a result, the 5th-order CPD model in (7) is partially symmetric:

$$\mathcal{T}(i_1, j_1, i_2, j_2, r) = \mathcal{T}^*(i_2, j_2, i_1, j_1, r) \quad (8)$$

Thus far, with equations (3) to (8) we have modeled the ICA-CPD problem as the CPD of a partially symmetric 5th-order CPD tensor $\mathcal{T}$.

## 3. PROPOSED ALGORITHM

We use matrix decomposition and joint diagonalization (JD) to identify the CPD model in (7). More exactly, we matricize $\mathcal{T}$ into $T \in \mathbb{C}^{I^2 \times J^2K}$ as follows:

$$T((i_1-1)I+i_2, (j_1-1)JK+(j_2-1)K+k) = \mathcal{T}(i_1, j_1, i_2, j_2, k) \quad (9)$$

By definition, $T$ could be written in the following form:

$$T = (A \odot A^*) \cdot (B \odot B^* \odot D)^T \quad (10)$$

In addition, by performing singular value decomposition (SVD): $T = U\Lambda V^H$, and comparing it with (10) we have:

$$\begin{cases} A \odot A^* = UF \\ B \odot B^* \odot D = V^*F^{-T} \end{cases} \quad (11)$$

where $U = [u_1, ..., u_R] \in \mathbb{C}^{I^2 \times R}$, $V = [v_1, ..., v_R] \in \mathbb{C}^{J^2K \times R}$, and $F$ is a $R$ by $R$ invertible matrix.

Next, we shall prove that $F$ is real-valued under partial symmetry of $\mathcal{T}$ which intuitively yields the following:

$$\begin{aligned} T((i_1-1)I+i_2, (j_1-1)JK+(j_2-1)K+k) \\ = T^*((i_2-1)I+i_1, (j_2-1)JK+(j_1-1)K+k) \end{aligned} \quad (12)$$

Substituting $T = U\Lambda V^H$ into (12) yields following result after several derivations:

$$U_r(i_1, i_2)\mathcal{V}_r^*(j_1, j_2, k) = U_r^*(i_2, i_1)\mathcal{V}_r(j_2, j_1, k) \quad (13)$$

where $U_r \in \mathbb{C}^{I \times I}$, and $\mathcal{V}_r \in \mathbb{C}^{J \times J \times K}$ are defined as follows:

$$\begin{cases} U_r(i_1, i_2) = u_r((i_1-1)I+i_2) \\ \mathcal{V}_r(j_1, j_2, k) = v_r(j_1-1)JK+(j_2-1)K+k) \end{cases} \quad (14)$$

Then we could prove the following theorem:

*Theorem 1:* There exists a unit-modulus scalar $\alpha$ such that $\alpha U_r$ and $\alpha^*\mathcal{V}_r(:,:,k)$ are both Hermitian if (13) holds (here we use matlab notation $\mathcal{V}_r(:,:,k)$ to denote the matrix obtained by fixing the third index of $\mathcal{V}_r$ to $k$).

The proof of above theorem is similar to that of Theorem 1 in [13]. The calculation of $\alpha$ could be found in [13] as well. In the following, we assume $U_r$ and $\mathcal{V}_r$ are already normalized by $\alpha$ to possess Hermitianity properties.

We rewrite the first equation of (11) as:

$$\sum_{r=1}^{R} U_r(i_1, i_2) \cdot F(r,u) = A(i_1,u) \cdot A^*(i_2,u) \quad (15)$$

Then we have the following by calculating the conjugate of (15) and taking into account the Hermitianity of $U_r$:

$$\sum_{r=1}^{R} U_r(i_2, i_1) \cdot F^*(r,u) = A(i_2,u) \cdot A^*(i_1,u) \quad (16)$$

---

[1] The sampled 4th-order cumulant matrix for $X, Y, Z, W \in \mathbb{C}^{I \times T}$ is given by:
$\text{cum}(X, Y, Z, W)_{(i-1)I+j,(k-1)I+l} \triangleq T^{-1}\sum_{t=1}^{T} x_{i,t} y_{j,t} z_{k,t} w_{l,t} - T^{-2}\sum_{t=1}^{T} x_{i,t} y_{j,t} \sum_{t=1}^{T} z_{k,t} w_{l,t}$
$-T^{-2} \cdot (\sum_{t=1}^{T} x_{i,t} z_{k,t} \sum_{t=1}^{T} y_{j,t} w_{l,t} + \sum_{t=1}^{T} x_{i,t} w_{l,t} \sum_{t=1}^{T} y_{j,t} z_{k,t})$

Comparing (15) and (16) we come to the conclusion that $F$ is real-valued. As a result, if we look back to (11), the CPD problem now amounts to finding the real-valued matrix $F$ such that $UF$ is of Khatri-Rao structure and $V^*F^{-T}$ is of double Khatri-Rao structure.

We borrow two rank-1 detecting tensors from [12] to solve the above problem, which are defined as:

$$\begin{cases} [\Phi_1(X,Y)]_{i,j,k,l} = x_{i,k}y_{j,l} + x_{j,l}y_{i,k} - x_{i,l}y_{j,k} - x_{j,k}y_{i,l} \\ [\Phi_2(\mathcal{X},\mathcal{Y})]_{i,j,k,l,m,n} = x_{i,k,m}y_{j,l,n} + x_{j,l,n}y_{i,k,m} - x_{j,k,m}y_{i,l,n} - x_{i,l,n}y_{j,k,m} \end{cases}$$
(17)

and use them upon $U_r$ and $\mathcal{V}_r$ to construct 2 tensors as:

$$\mathcal{P}_{r,u} = \Phi_1(U_r, U_u), \qquad \mathcal{Q}_{r,u} = \Phi_2(\mathcal{V}_r, \mathcal{V}_u) \quad (18)$$

We note that $\Phi_1(X,X) = \mathcal{O}$ iff $X$ is rank-1 ($\mathcal{O}$ denotes a tensor with all zeros), and $\Phi_2(\mathcal{X},\mathcal{X}) = \mathcal{O}$ iff the mode-1 matricization of $\mathcal{X}$, denoted by $X_1$ and defined by $X_1(i,(j-1)K+k) \triangleq \mathcal{X}(i,j,k)$, is rank-1. It is important to note that in our case there is no need to detect the rank-1 structure of the mode-2 matricization of $\mathcal{V}_r$, as is done in [12], mainly due to the Hermitianity of $\mathcal{V}_r(:,:,k)$.

As a result, with a few similar derivations to those in [12] given $\Phi_1(a_k a_k^H, a_r a_r^H)$ and $\Phi_2(b_k \circ b_k^* \circ d_k, b_r \circ b_r^* \circ d_r)$ are linearly independent for $k \neq r$, we conclude that there exist 2 sets of $R$ linearly independent complex matrices $M_r, W_r \in \mathbb{C}^{R \times R}$, $r = 1,\ldots,R$, such that:

$$\sum_{s,t=1}^{R} (M_r)_{s,t} \mathcal{P}_{s,t} = \mathcal{O}, \qquad \sum_{s,t=1}^{R} (W_r)_{s,t} \mathcal{Q}_{s,t} = \mathcal{O} \quad (19)$$

and $F, F^{-T}$ diagonalize these 2 sets of matrices, respectively:

$$\begin{cases} M_r = F\Sigma_r F^T \\ W_r = F^{-T}\Lambda_r F^{-1} \end{cases} \quad r = 1,2,\ldots,R \quad (20)$$

Noting that $W_r^{-1} = F\Lambda_r^{-1}F^T$ and $F$ is real-valued, $F$ could be finally obtained by performing real-valued non-orthogonal JD (RNJD) upon the union set of Re($M_r$), Im($M_r$), Re($W_r^{-1}$), and Im($W_r^{-1}$), $r = 1,2,\ldots,R$. Several options for RNJD are available in the open literature [14, 15].

When $F$ is estimated, we could calculate $A \odot A^*$ and $B \odot B^* \odot D$ via (11). Therefore, the loading matrices could be finally obtained with the rank-1 approximation based scheme upon $A \odot A^*$ and $B \odot B^* \odot D$. Details about rank-1 approximation could be found in [12]. We summarize the proposed 5th-order CPD with Partial Symmetry via JD (CPS5-JD) in TABLE I. We note that the proposed method calculates the loading matrices with 3 major mathematical tools including SVD, RNJD, and rank-1 approximation, which are all computa-tionally efficient and involve no updating iterations for the loading matrices. Therefore, CPS5-JD is expected to be faster than the ALS-ELS based CPD in [10]. Moreover, the proposed CPS5-JD algorithm, as far as we know, is the first JD based 5th-order CPD algorithm which takes into account partial symmetry, and this distincts itself from the methods in [13] that use super-symmetry of 4th-order tensors.

TABLE I.
FIFTH-ORDER CPD WITH PARTIAL SYMMETRY VIA JD (CPS5-JD)

| |
|---|
| **Input:** $\mathcal{T} \in \mathbb{C}^{I \times J \times I \times J \times R}$ with partial symmetry (8), and rank $R$ |
| **1:** Matricize $\mathcal{T}$ into $T$ via (9), and do SVD on $T$: $T = UV^H$; |
| **2:** Calculate tensors $\mathcal{P}_{r,u}$ and $\mathcal{Q}_{r,u}$, $r,u = 1,\ldots,R$ via (14), (17), and (18); |
| **3:** Obtain matrices $M_r, W_r$, $r = 1,2,\ldots,R$ by solving (19); |
| **4:** Calculate $F$ via RNJD upon the union set of Re($M_r$), Im($M_r$), Re($W_r^{-1}$), and Im($W_r^{-1}$), $r = 1,2,\ldots,R$; |
| **5:** Estimate $A, B, D$ from $UF$ and $V^*F^{-T}$ via rank-1 approximation; |
| **Output:** The loading matrices $A, B, D$ |

## 4. SIMULATIONS

In this section, we use numerical simulations to demonstrate the performance of the proposed algorithm. The proposed CPS5-JD algorithm is compared with 5th-order CPD with partial symmetry based on ELS and ALS (CPS5-EALS). We note here that CPS5-EALS is initialized with randomly generated orthonormal factor matrices. Computing configurations for performing the simulations are summarized as follows, CPU: Intel Core i7 2.93GHz; Memory: 16GB; System: 64bit Windows 7; Matlab R2010b.

**Simulation 1:** We construct a partially symmetric 5th-order tensor by (7) and (8). The loading matrices $A, B \in \mathbb{C}^{6 \times 5}$ are generated to incorporate highly collinear structures as follows: The $j$th and ($j$-1)th columns of $A$ are generated as: $a_j = a_{j-1} + 0.08v_j$, $j = 2,3,\ldots, 5$, and $a_1 = v_1$, with both the real and imaginary parts of $v_j$ drawn from standard normal distributions. The loading matrix $B$ is generated in the same way as $A$. The entries of the loading matrix $C \in \mathbb{R}^{6 \times 5}$ at the 5th mode are drawn from standard normal distributions. By definition we note here that the target tensor is of size $\mathcal{T} \in \mathbb{C}^{6 \times 6 \times 6 \times 6 \times 6}$, with the tensor rank $R = 5$. We add a noise term into the target tensor in the following way:

$$\tilde{\mathcal{T}} = \frac{\mathcal{T}}{\|\mathcal{T}\|_F} + \sigma \cdot \frac{\mathcal{N}}{\|\mathcal{N}\|_F} \quad (21)$$

where $\sigma$ denotes the noise level. Therefore, we could define the signal-to-noise ratio (SNR) as $snr = -10\log_{10}\sigma$. We evaluate the performance of all the compared algorithms by Amari's performance index (PI).

We let SNR vary from 20 – 80 dB, take 200 independent runs for each fixed SNR point, and draw the PI curves for both estimates of $A$ and $B$ for all the compared algorithms. The results are given in Figure 1. At the meantime, we collect in TABLE II the averaged running times for all the competitors as well.

From the results we could clearly see that the proposed CPS5-JD algorithm generates much faster computation than CPS5-EALS, as well as improved accuracy in the presence of high collinearities.

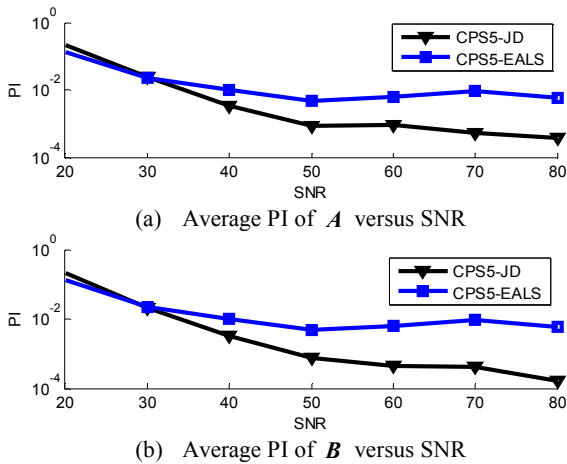

(a) Average PI of $A$ versus SNR

(b) Average PI of $B$ versus SNR

**Figure 1. Comparison of CPS5-JD and CPS5-EALS in the presence of high collinearities.**

TABLE II
RUNNING TIMES VERSUS SNR WITH HIGH COLLINEARITIES (IN SECOND)

| SNR(dB)<br>Methods | 20 | 30 | 40 | 50 | 60 |
|---|---|---|---|---|---|
| CPS5-JD | 0.2974 | 0.3496 | 0.3140 | 0.2985 | 0.2532 |
| CPS5-EALS | 0.9363 | 2.2213 | 2.7386 | 2.7578 | 2.1896 |

**Simulation 2:** We apply the proposed algorithm in ICA-CPD problem. More exactly, we construct three-way dataset $\mathcal{X} \in \mathbb{C}^{6 \times 5 \times 1000}$ following (1), The loading matrices $A \in \mathbb{C}^{6 \times 3}$ are generated to incorporate highly collinear structures as follows: The $j$th and $(j-1)$th columns of $A$ are generated as: $a_j = a_{j-1} + 0.1 v_j$, $j = 2,3$, and $a_1 = v_1$, with both the real and imaginary parts of $v_j$ drawn from standard normal distributions. The loading matrix $B$ is generated in the same way as $A$. 3 sources exist that are taken to be random phase signals. Gaussian white noises are added. SNR in this case is defined as $snr \triangleq 10\log_{10}(p_n^{-1} p_s)$, with $p_n$ and $p_s$ being the noise power and signal power, respectively. We perform ICA-CPD outlined in (3) to (8), with CPS5-JD and CPS5-EALS as candidates for the decomposition of the CPD model in (7).

We let SNR vary from 10 – 50 dB, take 200 independent runs for each fixed SNR points, and draw the PI curves for both estimates of $A$ and $B$ for all the compared algorithms. The results are given in Figure 2. Meanwhile, we collect the average running times for all the competitors in TABLE III.

From the results we observe that the proposed CPS5-JD yields much improved accuracy than CPS5-EALS in the presence of high collinearities. In addition, we could see from TABLE III that the running times for CPS5-JD across distinct SNR values are quite consistent, which are much less than CPS5-EALS (note here that the runtimes of CPS5-EALS at low SNR's make no sense as for these SNR values CPS5-EALS fails to generate correct results).

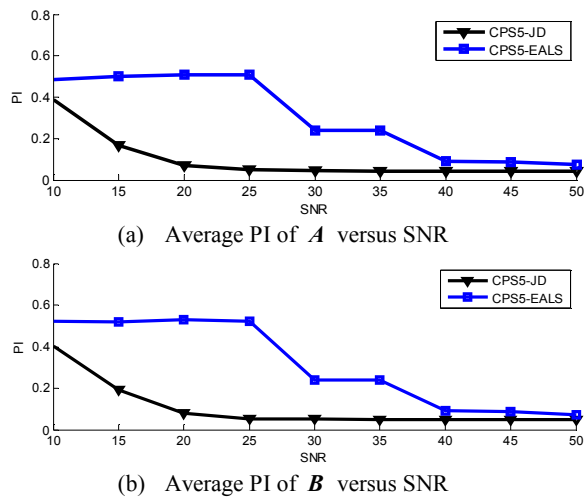

(a) Average PI of $A$ versus SNR

(b) Average PI of $B$ versus SNR

**Figure 2. Comparison of CPS5-JD, CPS5-EALS in ICA-CPD applications.**

TABLE III
RUNNING TIMES VERSUS SNR IN ICA-CPD (IN SECOND)

| SNR(dB)<br>Methods | 10 | 20 | 30 | 40 | 50 |
|---|---|---|---|---|---|
| CPS5-JD | 0.0413 | 0.0437 | 0.0432 | 0.0400 | 0.0281 |
| CPS5-EALS | 0.0020 | 0.0020 | 1.0597 | 1.2270 | 1.1543 |

## 5. CONCLUSION

This study presents a new 5th-order partially symmetric canonical polyadic decomposition (CPD) algorithm (CPS5-JD), for the problem of combined independent component analysis and canonical polyadic decomposition (ICA-CPD), via real-valued non-orthogonal joint diagonalization. Simulations have shown that the main merit of the proposed algorithm is the much improved computation speed over the existing method with alternating least squares and enhanced line search (CPS5-EALS) as well as improved accuracy, particularly in difficult situations with collinearities present.